\documentclass[10pt,conference]{IEEEtran}
\IEEEoverridecommandlockouts
\usepackage{verbatim}
\usepackage{cite}
\usepackage{amsmath,amssymb,amsfonts}
\usepackage{algorithmic}
\usepackage{graphicx}
\usepackage{textcomp}
\usepackage{xcolor}
\usepackage[table]{xcolor}
\usepackage{algorithm}

\usepackage{tabularx}
\usepackage{multirow}
\usepackage[table]{xcolor}

\usepackage{microtype}
\def\BibTeX{{\rm B\kern-.05em{\sc i\kern-.025em b}\kern-.08em
    T\kern-.1667em\lower.7ex\hbox{E}\kern-.125emX}}
\begin{document}

\title{Adaptive Multimodal Agents-Based Framework for Automatic Workflow Execution}


\author{\IEEEauthorblockN{Susanna Cifani}
\IEEEauthorblockA{
\textit{Sapienza University of Rome}\\
Rome, Italy 
}
\and
\IEEEauthorblockN{Mario Luca Bernardi}
\IEEEauthorblockA{\textit{Department of Engineering} \\
\textit{University of Sannio}\\
Benevento, Italy \\
bernardi@unisannio.it}
\and
\IEEEauthorblockN{Marta Cimitile}
\IEEEauthorblockA{\textit{Faculty of Jurisprudence} \\
\textit{Unitelma Sapienza University}\\
Rome, Italy \\
marta.cimitile@unitelmasapienza.it}
}

\maketitle

\begin{abstract}
Modern information systems require autonomous agents capable of navigating complex workflows, yet current methodologies often struggle with the transition from structured metadata parsing to general environmental perception. While the integration of MLLMs has enabled agents to interact directly with GUIs, existing approaches typically treat task sequences as discrete, linear episodes. This fragmentation prevents agents from capturing the underlying transition topology, limiting their effectiveness in novel or non-stationary scenarios. To address this, we propose a novel multimodal multi-agent framework that achieves automatic workflow execution through a distinct two-phase pipeline. First, during an offline discovery phase, the architecture adaptively constructs a topological knowledge base from fragmented execution logs. During inference, agents leverage Adaptive Retrieval-Augmented Generation (RAG) over this fixed, pre-established graph, coupled with a closed-loop collaborative verification protocol to dynamically self-correct and navigate. This graph-based approach facilitates superior task decomposition and adaptive navigation performance. We validate our framework in a real-world context, demonstrating its ability to maintain high reliability and semantic awareness even with limited training data.
\end{abstract}


\section{Introduction}

In the context of modern information systems, the autonomous execution of tasks within workflow-driven environments has emerged as a pivotal challenge for software engineering and human-computer interaction \cite{Liuetal2025,DBLP:journals/jiis/BernardiCCM24}. While early methodologies relied on parsing structural metadata to facilitate Large Language Model (LLM) decision-making \cite{yang2023mind2web, yin2023survey}, the integration of Multimodal Large Language Models (MLLMs) \cite{mishra2024multimodal} has shifted the paradigm toward agents capable of analyzing system states and executing transitions end-to-end \cite{cheng2024seeclick}. However, a significant limitation of existing approaches \cite{mishra2024multimodal, yang2023mind2web}  is the reliance on linear, discrete episodes, which treat task sequences as independent items and fail to capture the complex, interconnected nature of system-state transitions. 

This lack of holistic perception restricts an agent's ability to adapt to non-stationary interfaces or navigate novel workflows, as it ignores the multivariable action-result relationships that define the broader operational environment. Consequently, when faced with novel or complex workflows, an agent limited by isolated historical traces cannot synthesize its fragmented knowledge to discover new paths, whereas a graph-based perception allows it to leverage the underlying transition topology to anticipate the consequences of alternative actions and adaptively navigate toward a target state with far greater flexibility \cite{pan2024unifying,guan2024knowledgetopology,edge2024graphrag}.

Our approach relies on a workflow graph constructed during an offline discovery phase. To achieve this, the framework utilizes an automated pipeline for state-jump determination and node similarity checking, reconstructing fragmented execution logs into a coherent topological knowledge base. In this framework, historical task executions are interpreted as specific path samplings of this broader structure, allowing the system's global knowledge to incrementally grow and adapt to new observations prior to active inference.

During real-time execution, agents powered by LLMs navigate this established, fixed topology by leveraging Adaptive Retrieval-Augmented Generation (RAG) \cite{lewis2020rag} to extract graph-structured guidelines. To interact directly with the system's GUI without losing context, the agents utilize a dynamic, semantic history mechanism. Rather than merely recording atomic actions, this mechanism generates a goal-aware narrative of state transitions, ensuring continuous alignment with the overall objective and preventing infinite loops. Moving beyond traditional "open-loop" systems, we implement a multi-agent verification protocol acting as a closed-loop self-correction mechanism; in this collaborative environment, specialized agents intercept and validate the activities of their peers before any action is accepted as input for the next step. This verification stage ensures that only logically sound state transitions are integrated into the ongoing task execution. Ultimately, this approach enables superior semantic perception and task decomposition by providing agents with an authentic set of transition relationships, significantly enhancing navigation performance even when training data is limited.

 This study also proposes a validation of our approach in a real context. To this aim, we evaluate the effectiveness of the proposed architecture on the \textit{GUIOdyssey} \cite{lu2025} benchmark, a freely available dataset which comprises 1,666 episodes of cross-app mobile navigation tasks.
 We also perform a comparative analysis of the proposed approach effectiveness compared to the state-of-the-art approaches. The experiments show good performance of the proposed approach that  triples the Success Rate (SR) of the best alternative approach demonstrating its critical role in automatic workflow execution.

The structure of this paper is organized as follows: Section \ref{sec:related} provides an overview of existing literature, followed by Section \ref{sec:background}, which establishes the theoretical and technical foundations. The core methodology of our proposed workflow-based approach is detailed in Section \ref{sec:approach}. We then present the experimental framework and an analysis of the results in Section \ref{sec:exp} and Section \ref{sec:results}. The paper concludes with Section \ref{sec:con}, which offers a summary of our findings and a discussion of future research directions.

\section{Related Work}
\label{sec:related}

The autonomous execution of tasks within workflow-driven environments has become a focal point in software engineering and human-computer interaction \cite{Liuetal2025}. Current research can be broadly categorized into two main paradigms: structural-based and multimodal-based approaches.

Early methodologies in digital task automation focused on parsing structural metadata (such as HTML trees or API schemas) to guide LLM decision-making \cite{liu2024webllama}. A notable example is Mind2Web \cite{yang2023mind2web}, which introduced a benchmark for generalist web agents.
Furthermore, earlier frameworks often utilized reinforcement learning and supervised fine-tuning on structural traces, as seen in WebShop \cite{yao2022webshop}, which demonstrated how agents could learn to navigate e-commerce platforms by parsing product attributes and search results. 
Authors in \cite{IVERSEN2026} leverages a LLM for automated compliance checking (ACC) to directly interpret regulations.
However, these methods often struggleto adapt to dynamic, non-stationary interfaces where structural metadata is incomplete, inconsistent, or constantly evolving.
Moreover, they often fail when visual cues (such as icon-based buttons without text labels) are the primary means of navigation, a limitation that has paved the way for the multimodal approaches.
The evolution of MLLMs \cite{mishra2024multimodal} has shifted the paradigm toward agents that treat the Graphical User Interface (GUI) as a visual landscape. Recent frameworks like SeeClick \cite{cheng2024seeclick} leverage these models to interpret system states and execute transitions end-to-end. Despite their visual prowess, these systems frequently treat task sequences as discrete, linear episodes \cite{mishra2024multimodal, yang2023mind2web}. This fragmentation prevents the agent from understanding the underlying transition topology of the system.
In these frameworks \cite{mishra2024multimodal, yang2023mind2web}, each task is typically executed in isolation, where the agent perceives a sequence of states as a unique, non-reusable trace. For instance, models like GPT-4V or specialized agents such as CogAgent\cite{hong2023cogagent} demonstrate remarkable zero-shot capabilities in interpreting single GUI screenshots; however, they lack a persistent memory of the global system dynamics.

This "tunnel vision" is evident in benchmarks like VisualWebArena \cite{koh2024visualwebarena}, where agents are evaluated on their ability to complete tasks through direct visual interaction. While successful in simple paths, these agents fail to recognize that different tasks often share common system states, leading to redundant explorations and an inability to recover from errors. As noted in recent critiques of the "chain-of-action" paradigm \cite{zheng2024gpt4v}, the reliance on independent history traces prevents the agent from synthesizing a mental map of the environment. Without a graph-based representation, the agent cannot identify that state $s_i$ in Task A is functionally identical to state $s_j$ in Task B, effectively ignoring the underlying transition topology. Consequently, when faced with novel workflows, these linear agents cannot leverage previously seen "shortcut" transitions, whereas an adaptive topological approach allows for dynamic pathfinding across the entire operational manifold.
Authors in \cite{chen2025} introduce the PG-Agent framework  as a novel approach to GUI navigation by transitioning from traditional linear episode-based memory to a structured graph representation. This methodology addresses the  above discussed of sequential models, showing improved capability to capture complex transition relationships between different application states, thereby improving the agent's ability to generalize across unseen scenarios.
The proposed study introduces a framework comparable to the PG-Agent for the surrounding idea  of a structured graph representation of the system workflow.
Unlike the PG-Agent framework, which logs history through mechanical, rule-based descriptions, we introduce a goal-oriented semantic history. This narrative approach captures the underlying intent of each action and selectively extracts only the environmental information necessary to achieve the goal, providing agents with a goal-aware context for making more informed decisions.
To support this adaptive behavior, this study also proposes a closed-loop multi-agent verification protocol where specialized agents intercept and validate the activities of their peers before any action is performed.
This verification layer acts as a semantic gateway, filtering out inconsistent state transitions to ensure that the agent's ongoing execution trajectory maintains logical integrity.

\section{Background: Knowledge Graphs in RAG}
\label{sec:background}

The fundamental principle of the RAG framework resides in the synergistic integration of the expansive parametric knowledge of LLMs with specialized, high-fidelity, and dynamic data sourced from external repositories \cite{lewis2020rag}. Operationally, the process is initiated by an input query, which triggers a retrieval mechanism to identify and extract pertinent information from a corpus via sophisticated search algorithms \cite{gao2024retrievalaugmented}. This retrieved information is subsequently concatenated into the LLM's prompt, effectively utilizing In-Context Learning (ICL) \cite{dong2023} to provide the model with a grounded, non-parametric context. By combining these information retrieval mechanisms with generative capabilities, RAG significantly enhances the performance of LLMs, mitigating issues related to hallucinations and ensuring that outputs remain contextually relevant to the specific operational domain.
In the domain of Knowledge-Intensive Natural Language Processing, RAG has traditionally functioned by optimizing the conditional probability $P(y \mid q, D)$, where $D$ represents a set of unstructured document chunks retrieved from a dense vector space based on semantic proximity to a query $q$ \cite{lewis2020rag}. However, recent advancements have shifted this paradigm toward Graph-based RAG (GraphRAG), which formalizes the retrieval corpus as a structured knowledge graph $\mathcal{G} = (\mathcal{V}, \mathcal{E}, \mathcal{R})$, where $\mathcal{V}$ denotes entities or system states, $\mathcal{E}$ represents directed edges of interaction, and $\mathcal{R}$ signifies the semantic predicates defining their relations \cite{ji2021survey, pan2024unifying}. 

Unlike standard retrieval, which is often limited by the "lost in the middle" phenomenon and an inability to model multi-hop dependencies, GraphRAG enables the extraction of a relevant subgraph $\mathcal{G}' \subset \mathcal{G}$ to provide a topological context for LLM reasoning \cite{edge2024graphrag, yasunaga2022qagnn}. By linearizing triplets $(v_i, r_{ij}, v_j)$ into the prompt, the model can navigate complex operational manifolds, leveraging the underlying transition logic to synthesize decision-making paths that would remain obscured in fragmented, linear episodic traces \cite{guan2024knowledgetopology}. This structural grounding is essential for autonomous agents to achieve holistic perception in non-stationary environments, allowing them to anticipate the consequences of alternative actions, adapt to interface changes, and navigate toward target states by utilizing the inherent connectivity of the information system.
\begin{algorithm}[t]
\caption{Multi-Agent Workflow Execution}
\label{alg:execution}
\begin{algorithmic}[1]
\REQUIRE Task query $q$, Knowledge base $\mathcal{K}$, Max retries $M$
\ENSURE Task completion status
\STATE $\mathcal{T}_{\text{top-}k} \leftarrow \mathcal{R}(q, \mathcal{K})$ \COMMENT{Retrieve relevant traces}
\STATE $c \leftarrow \text{ContextBuilder}(\mathcal{T}_{\text{top-}k})$ \COMMENT{Build augmented context}
\STATE $\mathcal{G} \leftarrow \text{GlobalPlanner}(q, c)$ \COMMENT{Generate global plan}
\WHILE{task not complete}
    \STATE $g_t \leftarrow \text{SubgoalPlanner}(\mathcal{G}, h)$ \COMMENT{Get current sub-goal}
    \STATE $s_t \leftarrow \text{ObservationAgent}()$ \COMMENT{Encode current state}
    \STATE $m \leftarrow 0$ \COMMENT{Initialize retry counter}
    \REPEAT
        \STATE $a_t \leftarrow \text{DecisionAgent}(g_t, s_t)$ \COMMENT{Propose action}
        \STATE $(v, f) \leftarrow \text{VerifierAgent}(s_t, a_t, g_t)$ \COMMENT{Verify}
        \IF{$v = \textsc{Reject}$}
            \STATE $g_t \leftarrow \text{SubgoalPlanner}(g_t, f)$ \COMMENT{Refine with feedback}
            \STATE $m \leftarrow m + 1$
        \ENDIF
    \UNTIL{$v = \textsc{Approve}$ \OR $m \geq M$}
    \STATE Execute action $a_t$
    \STATE $h \leftarrow h \cup \{\text{Narrate}(s_t, a_t, s_{t+1})\}$ \COMMENT{Update history}
\ENDWHILE
\end{algorithmic}
\end{algorithm}

\section{The Multimodal Agent-based Framework}
\label{sec:approach}
\begin{figure*}[t]
    \centering \includegraphics[width=0.87\textwidth]{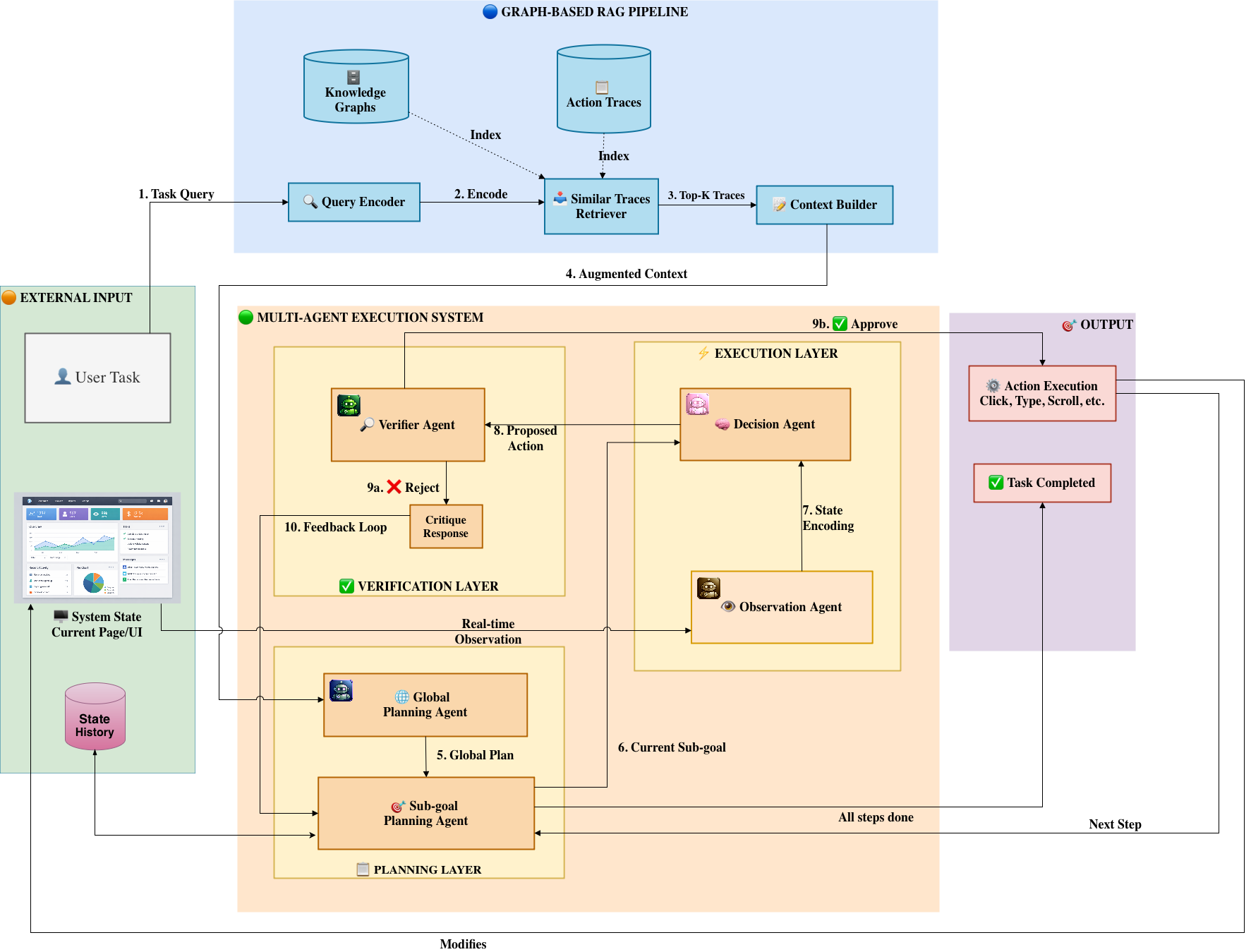}
\caption{Simplified schema of the proposed framework. }
\label{fig:workflow} 
\end{figure*}

The proposed framework, illustrated in Figure~\ref{fig:workflow}, comprises two principal components: an Adaptive Graph-based RAG Pipeline for knowledge augmentation and a Multi-Agent Execution System for task completion. The process begins when a user task is encoded to retrieve relevant knowledge from indexed Knowledge Graphs and Action Traces, providing agents with an Augmented Context that transcends the limitations of isolated, linear episodes.


Within the execution system, a hierarchical planning structure decomposes tasks while an Observation Agent captures real-time system state to inform action proposals. Crucially, the framework employs a Verification Layer where a Verifier Agent audits proposed transitions, either triggering a Feedback Loop for correction or granting Approval for execution. The overall execution procedure is formalized in Algorithm~\ref{alg:execution}.

The procedure begins with the {adaptive knowledge retrieval phase} (Lines 1--3): given a task query $q$, the system retrieves the top-$k$ most similar traces from the knowledge base $\mathcal{K}$ using a retrieval function $\mathcal{R}$, then constructs an augmented context $c$ that grounds subsequent planning.

The {planning phase} (Line 4) invokes the Global Planning Agent to decompose the query into a high-level strategy $\mathcal{G}$, which serves as a roadmap for the entire task execution.

The main execution loop (Lines 5--17) implements the {closed-loop action cycle}. At each iteration, the Sub-goal Planning Agent derives the current sub-goal $g_t$ from the global plan $\mathcal{G}$ and the accumulated history $h$ (Line 6). The Observation Agent then encodes the current GUI state $s_t$ into a semantic representation (Line 7).

The {verification loop} (Lines 8--14) constitutes the core self-correction mechanism. The Decision Agent proposes an action $a_t$ based on the current sub-goal and state (Line 10). The Verifier Agent evaluates this proposal according to the consistency predicate (Equation~\ref{eq:verifier}), returning either approval or rejection with constructive feedback $f$ (Line 11). Upon rejection, the Sub-goal Planning Agent refines its strategy using the feedback (Line 13), and the cycle repeats for up to $M$ attempts. This recursive refinement ensures that only logically consistent actions proceed to execution.

Upon approval (or exhaustion of retry attempts), the action executes on the system (Line 15). Finally, the {history update} (Line 16) invokes the Narrate function, which generates a differential state narrative capturing the semantic transition from $s_t$ to $s_{t+1}$ relative to the global goal. This narrative is appended to the history $h$, providing the Sub-goal Planning Agent with goal-aware context for subsequent iterations.

The algorithm terminates when the task completion condition is satisfied, as determined by the Global Planning Agent's assessment that all sub-goals have been achieved.
In the following subsections further details about the Rag Pipeline and the Multi-Agent Execution System are reported.

\subsection{The Graph-based RAG pipeline}

The RAG Pipeline constitutes the primary knowledge-augmentation stratum of the proposed framework, designed to alleviate the inherent parametric limitations of LLMs through the integration of external, domain-specific repositories \cite{lewis2020rag}. 


As illustrated in the architectural schema, the process is initiated by a \textit{Task Query} $q$, which is processed by a \textit{Query Encoder} to generate a high-dimensional semantic embedding. This embedding facilitates the operation of a \textit{Similar Traces Retriever}, which executes a similarity search across indexed \textit{Knowledge Graphs} $\mathcal{G} = (\mathcal{V}, \mathcal{E})$ and historical \textit{Action Traces} to identify the most relevant behavioral patterns. 


Formally, the pipeline optimizes:
\begin{equation}
\mathcal{R}(q, \mathcal{K}) \rightarrow \mathcal{T}_{\text{top-}k} = \underset{T \in \mathcal{K}}{\text{arg\,top-}k} \; \text{sim}(\mathbf{e}_q, \mathbf{e}_T)
\label{eq:retrieval}
\end{equation}
where $\mathcal{K}$ represents the global knowledge base, $\mathcal{T}$ denotes the retrieved trajectories, and $\text{sim}(\cdot, \cdot)$ computes the cosine similarity between embeddings.

The $k$ most relevant traces are processed by a Context Builder to synthesize an Augmented Context, leveraging ICL~\cite{dong2023} to ground the downstream Multi-Agent Execution System.


By merging structured topological guidelines with dynamic search algorithms \cite{gao2024retrievalaugmented}, this pipeline ensures that the agents possess a holistic perception of the system's state-transition manifold, significantly reducing the probability of logical hallucinations during task execution.

During the offline discovery phase, the incremental construction of the evolving workflow graph ($\mathcal{G} = (\mathcal{V}, \mathcal{E})$) is facilitated by an automated discovery pipeline that transforms raw sequential episodes into a directed topological structure where nodes $\mathcal{V}$ represent unique GUI states and edges $\mathcal{E}$ denote the transition-inducing actions. To optimize the computational efficiency of the \textit{Knowledge Graphs} repository, a stochastic sampling strategy is employed, processing a randomized subset (1/50th) of the total training episodes across all categories. This aligns with standard practices in LLM-based agent evaluation, where extensive computational overhead necessitates data subsampling (e.g., AITW \cite{rawles2023androidinthewild}). Following the methodology and empirical findings established by PG-Agent \cite{chen2025}, this sampling ratio provides an optimal trade-off: it fully preserves topological fidelity, ensuring computational tractability without significant performance degradation. This synthesis process, which serves as the foundational data for the \textit{Query Encoder} and \textit{Similar Traces Retriever}, is executed through three distinct stages.
In the first stage leveraging a MLLM, the system evaluates action tuples $(I_{before}, a, I_{after})$ to distinguish between in-page modifications and genuine state transitions. Only actions identified as triggering a "page jump" are utilized to instantiate new transitions within the graph.
In the following step, to ensure graph conciseness and prevent node redundancy, a dual-level similarity check is integrated into the \textit{Similar Traces Retriever} logic. This mechanism utilizes \textit{BGE-M3} \cite{chen2024m3} semantic embeddings in conjunction with visual comparisons to determine if a state should be mapped to an existing node or initialized as a new entity. The system employs \textit{FAISS} \cite{refahi2025fast} to index these embeddings, retrieving the top-4 candidate nodes for high-fidelity comparison, a value empirically found to balance retrieval recall with computational latency. 
In the final \textit{Graph Update} phase, new nodes and directed edges are inserted into the repository, completing the topological structure that serves as a fixed knowledge base during inference. To streamline the navigation manifold, consecutive in-page operations are condensed into single edges, thereby simplifying the topological path while preserving the functional trajectory toward the target goal.

\subsection{Multi-Agent Execution System}
\label{our_multiagent_workflow}
The Multi-Agent Execution System is organized into three specialized layers that interact to transform high-level natural language instructions into precise GUI actions, ensuring robustness through a continuous validation cycle.


\subsubsection{Planning Layer}
The Planning Layer is responsible for hierarchical task decomposition. It begins with the Global Planning Agent, which generates a high-level strategy (\textit{Step 5}). The Sub-goal Planning Agent breaks down this strategy into immediate objectives (\textit{Step 6}). Unlike traditional static planners, this layer is dynamic: it receives feedback from the Verification Layer (\textit{Step 10}), allowing it to refine or pivot the current sub-goal if the previous attempt was deemed unsuccessful or logically inconsistent, thereby adapting to unexpected or non-stationary interface states. This ensures that the agent does not just follow a rigid sequence, but adapts its reasoning to the evolving state of the task.
The Sub-goal Planning Agent defines the sub-goals also on the basis of states history.
This allows to avoid the so called context loss. This phenomenon occurs when the agent loses track of previously completed steps, frequently leading to action Hallucination. 
In such cases, the agent tends to re-execute past actions, either because it fails to recognize their successful completion or because it has discarded information critical to the task's next phase. To mitigate this issue, we implement a refined adaptive history mechanism that moves beyond simple atomic descriptions (e.g., "Scroll down"). Instead, our system processes the transition between the previous and current visual states in relation to the global goal. This allows the model to generate a differential state narrative that captures the evolution of the task. For instance, rather than a generic action label, the history records: "Scrolled down to reveal the complete song lyrics, which are...". This semantic enrichment is essential for complex workflows, such as cross-app translation, where specific data must be extracted and carried over to a different environment. By maintaining this narrative, the agent can explicitly recognize when it has already obtained the information necessary for the goal, signaling that it must proceed to the next stage of the plan. This approach directly addresses the problem of infinite Loops, an error particularly critical in datasets like GUIOdyssey. In cross-app navigation, the agent must maintain a clear awareness of having extracted data from one application before transitioning to the next; without an informative history, the system often falls into a loop, navigating backward indefinitely in a futile attempt to retrieve information it has already seen but failed to internalize. By transforming raw actions into a narrative of state progress, our architecture ensures the agent maintains a consistent trajectory toward task completion.

\subsubsection{Execution Layer}
The Execution Layer translates abstract sub-goals into concrete interactions. The Observation Agent performs a state encoding (\textit{Step 7}), transforming the current page into a semantic representation. This encoding identifies interactive elements and layout constraints that are critical for grounding the agent's reasoning in the current environment. The Decision Agent integrates the current sub-goal (\textit{Step 6}) from the Planning Layer with the state encoding provided by the Observation Agent and formulates an atomic action proposal (\textit{Step 8}), but does not trigger the execution immediately. Instead, the proposed action is treated as a hypothesis and passed to the Verification Layer for validation.

\subsubsection{Verification Layer}

A significant limitation of standard agentic paradigms is their open-loop nature: once an action is planned, it executes without intermediate validation. This exposes systems to the overconfidence phenomenon typical of Vision-Language Models, where incorrect actions trigger error propagation---a single mistake undermining all subsequent actions.
To overcome this limitation, our architecture transitions to a closed-loop system capable of online adaptation by introducing the Verifier Agent. This module acts as a critical oversight component operating between the planning and execution phases. Instead of a linear flow, our system implements a self-correction cycle: the Verifier analyzes the context and, if it identifies an inconsistency (such as typing into a search bar before it has been activated), it issues a "retry" verdict (\textit{Step 9a}) and generates constructive feedback (e.g., \textit{"Cannot type: the search bar is not active and the keyboard is not visible"}), which triggers a recursive refinement cycle. This feedback is passed back to the Sub-goal Planning Agent (\textit{Step 10}), which revises the strategy and then the Decision Agent reformulates the action based on this corrected guidance. This internal loop can repeat for up to four attempts per step. If the Verifier Agent analyzes the atomic action and declares it correct, it proceeds to approve it (\textit{Step 9b}).

Given a proposed action $a_t$ in state $s_t$, the Verifier evaluates the logical consistency of the transition $(s_t, a_t)$ with respect to the current sub-goal $g_t$:
\begin{equation}
v(s_t, a_t, g_t) = 
\begin{cases}
\textsc{Approve} & \text{if } \mathcal{C}(s_t, a_t, g_t) = \text{True} \\
\textsc{Reject}(f) & \text{otherwise}
\end{cases}
\label{eq:verifier}
\end{equation}

where $\mathcal{C}(\cdot)$ denotes a consistency predicate and $f$ represents constructive feedback (e.g., ``Cannot type: search bar inactive, keyboard not visible''). Upon rejection, feedback propagates to the Sub-goal Planning Agent, which revises strategy before the Decision Agent reformulates the action. This recursive refinement cycle permits up to $M = 4$ attempts per step.

\section{Experimental Setup}
\label{sec:exp}
In this section, we detail the experimental setting designed to evaluate the effectiveness of our proposed architecture. 
In particular, in the following subsections respectively the description of the research questions and of the adopted dataset are reported.

\subsection{Research Questions}
Our evaluation is driven by two core research questions:
\begin{itemize}

\item \textbf{RQ1:} To what extent the proposed Multimodal Agent-based framework is effective in automatic workflow execution with respect to the SOTA method?
    

\item \textbf{RQ2:} What is the contribution of the integration to the overall performance of informative temporal context and self-correction mechanisms?
\end{itemize}

\subsection{Evaluation Metrics}

To address RQ1, we employ two primary metrics following Deng et al.~\cite{deng2023mind2web}:

\textbf{Action Matching Score (AMS).} This is a fine-grained metric quantifying the precision of the Decision Agent's output at each discrete step $t$:
\begin{equation}
\text{AMS} = \frac{1}{N} \sum_{i=1}^{N} \frac{1}{T_i} \sum_{t=1}^{T_i} \mathbb{1}[a_t^{(i)} = a_t^{*(i)}]
\end{equation}
where $a_t^{(i)}$ denotes the predicted action and $a_t^{*(i)}$ the ground-truth optimal action derived from the knowledge graph for episode $i$ with $T_i$ steps.

\textbf{Success Rate (SR).} It is a combined metric assessing complete task achievement:
\begin{equation}
\text{SR} = \frac{N_s}{N_{\text{total}}}
\end{equation}
where $N_s$ denotes successfully completed tasks and $N_{\text{total}}$ the total number of episodes. SR is a significantly more stringent metric than AMS due to its binary nature: even a single incorrect step results in a failed episode.

\vspace{-2mm}
\subsection{Dataset}
We conduct our experiments on the GUIOdyssey benchmark, utilizing the full test set, which comprises 1,666 episodes of cross-app mobile navigation tasks organized in the following functional domains:
\begin{itemize}
 \item \textbf{Information Management:} Focuses on searching, organizing and recording data across apps, like finding a fact on a browser and saving it to a notes application.
 \item \textbf{Web Shopping:} Involves the full lifecycle of online commerce, including product research, price comparison across different platforms, and the final purchase. 
 \item \textbf{Media Entertainment:} Covers activities within video and music streaming services (e.g., YouTube, Spotify, Netflix), including content discovery and playlist management. 
 \item \textbf{Social Sharing:} Tasks centered on content distribution, such as capturing a photo with the camera, editing it in a specialized app, and sharing the result across multiple social media platforms (e.g., Instagram, Threads). 
 \item \textbf{Multi-Apps:} The most complex category, involving high-level goals that necessitate fluid transitions across multiple domains and app combinations (e.g., organizing a movie night by coordinating film selection, snack orders, and digital invitations). 
\end{itemize}

\subsection{Computational Setup}
Experiments were conducted on 4 $\times$ NVIDIA A100 (40GB) GPUs using the vLLM framework with tensor parallelism. Processing all 1,666 GUIOdyssey episodes took $\sim$80 hours ($\sim$173s/episode). Per-step context lengths ranged from 4,000 to 8,300 tokens, with an average of 6,200 tokens.


\section{Results}
\label{sec:results}
In this section, we present and discuss the obtained results for each research question (as reported in the following subsections).

\subsection{Answer to RQ1: Comparison with State-of-the-Art}

Referring to the RQ1, Table \ref{tab:main_results} reports the AMS score across the different GUIOdyssey functional domains for different models.
The proposed Multimodal Agent-based framework shows improved score with respect to SOTA approaches for all the functional domains with the exception of the Information where PG-Agent (Qwen/72B) gives better results \cite{wang2024qwen2vl}.
Comparing PG-Agent (Qwen/72B)  and PG-Agent (Qwen/30B) we observed that, while the performance in categories like \textit{Information} and \textit{Shopping} significantly decreased, the 30B model showed competitive or even superior results in \textit{Tool} and \textit{Media}. This suggests that while model scale is critical for complex information extraction, the 30B model backbone remains a robust foundation for general navigation tasks. For this reason we adopt in our proposed approach a Qwen/30B model. Furthermore, evaluating the 72B model within our framework was computationally prohibitive: its weights ($\sim$144GB) leave insufficient VRAM on our cluster for the BGE embedding model and vLLM's KV cache required by our long contexts.
Table \ref{tab:main_results} shows that the higher AMS  evaluated across the functional domain is obtained when the proposed Multimodal Agent-based framework is used.
Our architecture achieved an overall AMS of \textbf{56.41\%}, marking an absolute improvement of \textbf{+12.85\%}.
The most significant finding is that our architecture with a 30B model outperforms the PG-Agent framework with a 72B model by \textbf{8.7} percentage points: this suggests the effective management of temporal context and closed-loop self-correction, which our History Construction and Verifier modules directly address.

\begin{table*}[t]
\centering 
\caption{Comparison of Action Matching Score (AMS) and Success Rate (SR) across different GUIOdyssey scenarios.}
\label{tab:main_results}
\small 
\setlength{\tabcolsep}{3pt} 
\begin{tabular}{|l|cc|cc|cc|cc|cc|cc|cc|} 
\hline
\multirow{2}{*}{\textbf{Method}} & \multicolumn{2}{c|}{\textbf{Tool}} & \multicolumn{2}{c|}{\textbf{Information}} & \multicolumn{2}{c|}{\textbf{Shopping}} & \multicolumn{2}{c|}{\textbf{Media}} & \multicolumn{2}{c|}{\textbf{Social}} & \multicolumn{2}{c|}{\textbf{Multi-Apps}} & \multicolumn{2}{c|}{\textbf{Overall}} \\
\cline{2-15}
 & \textbf{AMS} & \textbf{SR} & \textbf{AMS} & \textbf{SR} & \textbf{AMS} & \textbf{SR} & \textbf{AMS} & \textbf{SR} & \textbf{AMS} & \textbf{SR} & \textbf{AMS} & \textbf{SR} & \textbf{AMS} & \textbf{SR} \\
\hline
\rowcolor{blue!10} GeminiProVision & 3.3 & - & 4.0 & - & 2.3 & - & 4.3 & - & 1.5 & - & 3.2 & - & 4.9 & - \\
\rowcolor{blue!10} CogAgent & 11.8 & - & 15.7 & - & 10.7 & - & 9.2 & - & 11.7 & - & 13.1 & - & 10.7 & - \\
\rowcolor{blue!10} GPT-4V & 18.8 & - & 23.5 & - & 20.2 & - & 19.2 & - & 16.9 & - & 13.8 & - & 19.0 & - \\
\rowcolor{blue!10} Qwen2.5-VL-72B & 46.6 & - & 60.0 & - & 44.0 & - & 32.4 & - & 46.1 & - & 54.6 & - & 42.4 & - \\
\rowcolor{blue!10} PG-Agent (w/ 72B) & 48.6 & - & \textbf{61.5} & - & 47.2 & - & 35.5 & - & 46.9 & - & 52.6 & - & 47.7 & - \\
\hline
\rowcolor{yellow!20} PG-Agent (w/ 30B) & 53.5 & 0.84 & 38.8 & 0.0 & 35.6 & 0.0 & 39.0 & \textbf{0.89} & 47.9 & 0.83 & 43.7 & 0.0 & 43.6 & 0.42 \\
\rowcolor{yellow!20} \textbf{Proposed (w/ 30B)} & \textbf{68.2} & \textbf{3.90} & 52.8 & 0.0 & \textbf{49.5} & \textbf{1.16} & \textbf{50.8} & 0.48 & \textbf{58.9} & \textbf{1.32} & \textbf{55.2} & \textbf{1.54} & \textbf{56.4} & \textbf{1.55} \\
\hline
\end{tabular}
\end{table*}

Beyond the AMS, we observed improvements in task completion. It is important to note that the SR is a significantly more stringent metric than AMS due to its binary nature: in multi-app navigation tasks, SR is highly sensitive to error, as even a single incorrect step results in a failed episode. 
In highly complex cross-app environments, absolute SRs remain critically low across the literature (e.g., AITW \cite{rawles2023androidinthewild}, Mind2Web \cite{deng2023mind2web}, OSWorld \cite{xie2024osworld}, GUIOdyssey \cite{lu2025}). For this reason, following the established methodology of SOTA baselines like PG-Agent \cite{chen2025} (which completely omits SR) we primarily rely on the AMS. Nevertheless, our architecture more than tripled the baseline SR, increasing from 0.42\% in PG-Agent to \textbf{1.55\%} in the proposed approach (see Table \ref{tab:main_results}). The most significant result is observed in the challenging Multi-Apps category, where the PG-Agent fails to complete a single episode (0.0\% SR), whereas our proposed architecture achieves a 1.54\% SR.

To further validate the proposed framework, we extend our evaluation to include alternative model backbones. Due to computational constraints, this ablation was performed on a representative subset of 16 episodes rather than the full test set. As reported in Table \ref{tab:models_results}, we compared our 30B backbone against Qwen3-VL-32B-Thinking and GLM-4.6V-Flash. The results indicate that the Qwen3-VL-32B-Thinking model achieves the highest overall performance on this subset, reaching an AMS of \textbf{61.67\%} and a SR of \textbf{6.67\%}. While the GLM-4.6V-Flash model shows a lower overall AMS of \textbf{46.38\%} and fails to complete any episodes (0.0\% SR).

\begin{table*}[t]
\centering 
\caption{Comparison of Action Matching Score (AMS) and Success Rate (SR) across different GUIOdyssey scenarios.}
\label{tab:ablation_results}
\small 
\setlength{\tabcolsep}{3pt} 
\begin{tabular}{|l|cc|cc|cc|cc|cc|cc|cc|} 
\hline
\multirow{2}{*}{\textbf{Method}} & \multicolumn{2}{c|}{\textbf{Tool}} & \multicolumn{2}{c|}{\textbf{Information}} & \multicolumn{2}{c|}{\textbf{Shopping}} & \multicolumn{2}{c|}{\textbf{Media}} & \multicolumn{2}{c|}{\textbf{Social}} & \multicolumn{2}{c|}{\textbf{Multi-Apps}} & \multicolumn{2}{c|}{\textbf{Overall}} \\
\cline{2-15}
 & \textbf{AMS} & \textbf{SR} & \textbf{AMS} & \textbf{SR} & \textbf{AMS} & \textbf{SR} & \textbf{AMS} & \textbf{SR} & \textbf{AMS} & \textbf{SR} & \textbf{AMS} & \textbf{SR} & \textbf{AMS} & \textbf{SR} \\
\hline
\rowcolor{red!10} Temporal Context + Self-correction mechanism & 60.15 & 4.76 & 50.32 & 0.0 & 47.66 & 0.0 & 52.43 & 0.0 & 60.1 & 0.0 & 51.51 & 0.0 & 53.65 & 1.0 \\
\rowcolor{red!10} Temporal Context & \textbf{66.42} & 4.76 & \textbf{60.26} & 0.0 & \textbf{54.21} & 0.0 & \textbf{55.34} & 0.0 & \textbf{61.66} & 0.0 & \textbf{58.61} & 0.0 & \textbf{60.17} & 1.0 \\
\rowcolor{red!10} Self-correction mechanism & 63.47 & 4.76 & 48.08 & 0.0 & 43.93 & 0.0 & 52.43 & 0.0 & 58.55 & 0.0 & 52.55 & 0.0 & 53.72 & 1.0 \\
\hline
\end{tabular}
\end{table*}

\begin{table*}[t]
\centering 
\caption{Comparison of Action Matching Score (AMS) and Success Rate (SR) across different GUIOdyssey scenarios.}
\label{tab:models_results}
\small 
\setlength{\tabcolsep}{3pt} 
\begin{tabular}{|l|cc|cc|cc|cc|cc|cc|cc|} 
\hline
\multirow{2}{*}{\textbf{Method}} & \multicolumn{2}{c|}{\textbf{Tool}} & \multicolumn{2}{c|}{\textbf{Information}} & \multicolumn{2}{c|}{\textbf{Shopping}} & \multicolumn{2}{c|}{\textbf{Media}} & \multicolumn{2}{c|}{\textbf{Social}} & \multicolumn{2}{c|}{\textbf{Multi-Apps}} & \multicolumn{2}{c|}{\textbf{Overall}} \\
\cline{2-15}
 & \textbf{AMS} & \textbf{SR} & \textbf{AMS} & \textbf{SR} & \textbf{AMS} & \textbf{SR} & \textbf{AMS} & \textbf{SR} & \textbf{AMS} & \textbf{SR} & \textbf{AMS} & \textbf{SR} & \textbf{AMS} & \textbf{SR} \\
\hline
\rowcolor{green!10} Proposed (w/ 30B) & 60.0 & 0.0 & \textbf{66.67} & 0.0 & 72.73 & 0.0 & 33.33 & 0.0 & 60.38 & 0.0 & 37.7 & 25.0 & 54.05 & 6.25 \\
\rowcolor{green!10} Proposed (w/ Qwen3-VL-32B-Thinking) & \textbf{66.67} & 0.0 & 29.41 & 0.0 & \textbf{100.0} & 0.0 & \textbf{60.0} & \textbf{100.0} & \textbf{72.34} & 0.0 & \textbf{50.0} & 0.0 & \textbf{61.67} & \textbf{6.67} \\
\rowcolor{green!10} Proposed (w/ GLM-4.6V-Flash) & 50.0 & 0.0 & 23.53 & 0.0 & 72.73 & 0.0 & 44.44 & 0.0 & 60.38 & 0.0 & 34.67 & 0.0 & 46.38 & 0.0 \\
\hline
\end{tabular}
\end{table*}

\subsection{Answer to RQ2: Ablation Study}

To isolate the contributions of individual components, we conducted an ablation study on randomly sampled episodes $N = 100$, balancing statistical power against computational feasibility. We compare three configurations:
\begin{itemize}
    \item \textit{Full}: Both informative history and Verifier Agent
    \item \textit{Context-only}: Informative history without Verifier Agent
    \item \textit{Verifier-only}: Verifier Agent with non-informative history
\end{itemize}

The results in Table~\ref{tab:ablation_results} reveal a substantial disparity between the contributions of the components. The informative history mechanism provides the most significant performance improvement, increasing overall AMS from 53.65\% to 60.17\% (+6.52 pp). This suggests that maintaining contextual awareness is crucial to navigate complex GUI trajectories.

Conversely, the Verifier Agent in isolation yields marginal improvements (+0.07 pp overall AMS) and, in the \emph{Information} and \emph{Shopping} domains, produces slight performance degradation compared to the baseline. We hypothesize that without a solid contextual foundation, the self-correction mechanism struggles to adapt and identify accurate recovery strategies---the Verifier can detect inconsistencies but lacks the semantic grounding necessary to suggest productive alternatives.

In particular, the full architecture (53.65\% AMS) underperforms the context-only variant (60.17\% AMS) in terms of AMS. This counterintuitive result suggests potential interference: the Verifier's rejection cycles may disrupt otherwise correct action sequences when operating without sufficient confidence. Future work should investigate adaptive verification thresholds that modulate intervention based on confidence estimates.

To understand failure modes, we manually examined 50 failed episodes. The major failure categories are:
\begin{itemize}
    \item {Visual grounding errors} (38\%): Incorrect element localization, particularly for small or visually similar UI components.
    \item {Cross-app state loss} (26\%): Failure to maintain extracted information across application boundaries despite the semantic history mechanism.
    \item {Dynamic content handling} (22\%): Inability to adapt when UI elements load asynchronously or change position.
    \item {Verifier false positives} (14\%): Correct actions rejected due to overly conservative consistency checks.
\end{itemize}

These findings suggest that visual grounding remains the primary bottleneck, motivating future integration of specialized GUI grounding models.

\section{Conclusion and Future Work}

\label{sec:con}
This paper presented a multimodal multi-agent framework designed to overcome the limitations of linear episodic memory in autonomous workflow execution and adapt to dynamic interfaces. By integrating an Adaptive Graph-based RAG pipeline, the architecture reconstructs fragmented execution logs into a coherent knowledge base, enabling agents to leverage topological system-state transitions rather than isolated history traces. The introduction of a collaborative verification protocol transitions the system from an open-loop to a closed-loop paradigm capable of continuous self-correction, reducing logical hallucinations and error propagation.

Experimental results on the GUIOdyssey benchmark demonstrate substantial improvements: +12.8 percentage points in AMS and more than triple the SR compared to the strongest baseline. Notably, our 30B-parameter instantiation outperforms a 72B-parameter baseline, highlighting that architectural innovations---semantic history and closed-loop verification mechanisms---can exceed the benefits of raw model scaling.

Several limitations warrant acknowledgment. First, the overall SR (1.55\%), while substantially improved, remains low for practical deployment, indicating that GUI navigation remains a challenging open problem. Second, the ablation study reveals that our Verifier Agent provides marginal benefits in isolation and may interfere with the history mechanism, suggesting the need for more sophisticated context-aware integration strategies. Third, computational requirements for incremental graph construction and multi-agent inference limit scalability to very large action spaces. Fourth, evaluation on a single benchmark may not generalize to all GUI navigation scenarios.

We identify three priority directions: (1) integrating specialized visual grounding modules to address the dominant failure mode, (2) developing adaptive verification thresholds that modulate intervention based on action confidence, and (3) scaling the evolving graph-based representation to incorporate hierarchical abstraction for larger operational environments.

\section*{Acknowledgment}
We acknowledge financial support under the National Recovery and Resilience Plan (NRRP), M4C2I1.1, funded by the European Union – NextGenerationEU– Project Title aRtificial intElligence for Process Analytics (REPA) - Grant Assignment Decree No. 2022CJWPNA  (CUP UNITELMA  I53C24002420006) by the Italian Ministry of University and Research (MUR).

\bibliographystyle{plain}
\bibliography{biblio.bib}

\end{document}